# Adaptive network reliability analysis: Methodology and applications to power grid

Nariman L. Dehghani[1], Soroush Zamanian[1], and Abdollah Shafieezadeh[1,*]

[1] Risk Assessment and Management of Structural and Infrastructure Systems (RAMSIS) Lab, Department of Civil, Environmental, and Geodetic Engineering, The Ohio State University, Columbus, OH, 43210, United States

**ABSTRACT**

Flow network models can capture the underlying physics and operational constraints of many networked systems including the power grid and transportation and water networks. However, analyzing systems' reliability using computationally expensive flow-based models faces substantial challenges, especially for rare events. Existing actively trained meta-models, which present a new promising direction in reliability analysis, are not applicable to networks due to the inability of these methods to handle high-dimensional problems as well as discrete or mixed variable inputs. This study presents the first adaptive surrogate-based Network Reliability Analysis using Bayesian Additive Regression Trees ($ANR\text{-}BART$). This approach integrates $BART$ and Monte Carlo simulation ($MCS$) via an active learning method that identifies the most valuable training samples based on the credible intervals derived by $BART$ over the space of predictor variables as well as the proximity of the points to the estimated limit state. Benchmark power grids including IEEE 30, 57, 118, and 300-bus systems and their power flow models for cascading failure analysis are considered to investigate $ANR\text{-}BART$, $MCS$, subset simulation, and passively-trained optimal deep neural networks and $BART$. Results indicate that $ANR\text{-}BART$ is robust and yields accurate estimates of network failure probability, while significantly reducing the computational cost of reliability analysis.

* Corresponding author; e-mail address: shafieezadeh.1@osu.edu

# 1. INTRODUCTION

Societies increasingly rely on the performance of their lifeline networks such as the power grid and transportation and water networks. In fact, any disruptions in the operations of critical lifeline networks can lead to severe economic losses and incur significant hardship to communities (Paredes et al., 2019; Dehghani et al., 2021). These high consequences imparted by low probability events such as earthquakes and hurricanes in parallel with other factors, particularly, aging and population growth, demand for reliable risk management frameworks. Risk assessment and management of lifeline networks are key for planning and resource allocation decisions, which are among critical steps for building resilient communities (The White House, Office of the Press Secretary, 2016). An essential component of risk assessment and management of lifeline networks is network reliability analysis (Hardy et al., 2007; Song and Ok, 2010).

The efforts on reliability analysis of lifeline networks can be generally classified into two categories of connectivity reliability and flow network reliability (Duenas-Osorio and Hernandez-Fajardo, 2008). The connectivity analysis is concerned with whether there are connection paths between source nodes and terminal nodes. Recent efforts on network connectivity reliability analysis can be generally categorized into simulation-based and non-simulation-based system reliability approaches (Kim and Kang, 2013). Non-simulation-based approaches (*e.g.*, Li and He, 2002; Hardy et al., 2007; Kang et al., 2008), while can be applied to small size networks, are not able to handle large scale networks (Paredes et al., 2019). On the other hand, simulation-based approaches (*e.g.*, Adachi & Ellingwood, 2008; Nabian & Meidani, 2017; Dong et al., 2020) can estimate reliability of networks with complex and large-scale topologies. Although the connectivity reliability analysis provides the probability of having connection paths between source nodes and terminal nodes, it does not consider the extent of flow perturbation in networks, and consequently cannot properly capture the flow demand and capacity of networks (Chen et al., 1999; Duenas-Osorio and Hernandez-Fajardo, 2008). Alternatively, the flow network reliability analyzes the ability of networks in providing quality services. In this case, a network fails if the demand exceeds the capacity of the network. A detailed flow-based model involves solving flow optimization problems which is a computationally demanding process (Duenas-Osorio and Hernandez-Fajardo, 2008; Guidotti et al., 2017). The computational demand significantly increases as the size and complexity of networks increase.

Typical flow network reliability analysis relies on Monte Carlo simulation (*MCS*) (*e.g.,* Chen et al., 1999; Duenas-Osorio and Hernandez-Fajardo, 2008; Li et al., 2018). However, to properly approximate network reliability using the crude *MCS* approach, a large number of simulations is required, which adds to the computational complexity of the flow network reliability evaluation. Several sampling approaches, such as subset simulation (*SS*) (*e.g.,* Au and Beck, 2001) and Importance Sampling (*e.g.,* Echard et al., 2013) have been used to improve the efficiency of reliability estimation. While these techniques reduce computational costs relative to the *MCS* approach, the required number of simulations is often very large as they lack strategic selection of training samples (Wang & Shafieezadeh, 2019a). Moreover, the success of these methods in accurately and efficiently performing reliability analyses is not guaranteed (Papadopoulos et al., 2012). More specifically, *SS* exhibits large variability in estimation of the probability of failure as the accuracy of *SS* predictions significantly relies on an arbitrary selected set of parameters, when a priori knowledge about the optimal value of parameters is not available (Papadopoulos et al., 2012).

Recently, meta-models have offered a new direction for efficient reliability estimation. In these techniques, the original time-consuming model is emulated by a surrogate model that can be developed using a relatively small number of training samples (Bourinet, 2016; Gaspar et al., 2017). Several meta-modeling techniques, such as polynomial response surface, support vector machine, and neural networks have been used for this purpose. Although using these surrogate techniques can accelerate reliability analysis, in their passive form, they still require a large number of simulations to train the surrogate models. This is due to the inability of these techniques to prioritize samples with higher importance in the training process. Active learning reliability methods, on the other hand, can construct surrogate models for reliability analysis of systems using a limited number of Monte Carlo simulations. Among such techniques, methods that have integrated Kriging and *MCS* such as *AK-MCS* (Echard et al., 2011), *ISKRA* (Wen et al., 2016), *REAK* (Wang & Shafieezadeh, 2019b), and *ESC* (Wang & Shafieezadeh, 2019a) have gained wide attention. A significant advantage of Kriging is its ability to provide estimates for both the expected value and variance of responses over the space of inputs. This feature of Kriging has been used to identify the important training samples for adaptive training of the surrogate models. However, as the dimension of problems increases the performance of active learning reliability techniques based on Kriging degrades substantially (Zhang et al., 2018; Wang & Shafieezadeh, 2019a; Rahimi et al., 2020). Thus, the application of these methods is limited to reliability analysis of systems with a small number of components. Moreover, most active learning methods such as those based on Kriging are designed for continuous inputs. However, reliability analysis of networks commonly involves discrete variables (Chen et al., 2002). In fact, in network reliability analysis, random inputs are the states of a network's components, which are often considered as discrete random variables.

To address these limitations, this paper proposes an active learning approach for flow network reliability analysis of networked systems through integration of Bayesian Additive Regression Trees ($BART$) and $MCS$. To the best of authors knowledge, this paper presents the first adaptive surrogate-based reliability analysis for networks, in general. $BART$ is an advanced statistical model that is able to conform to high-dimensional data and capture highly nonlinear associations between predictor and dependent variables (Chipman et al., 2010). $BART$ also has a high predictive performance for data sets with categorical predictors and provides the best estimate of the dependent variables and the associated credible interval. Leveraging these features, this paper develops an approach for network reliability analysis called Adaptive Network Reliability analysis using $BART$ ($ANR$-$BART$). This approach first generates a Monte Carlo population in the design space. Each sample in this population is a binary vector with the size of the total number of vulnerable components in the network. A probabilistic criterion is proposed to determine the adequate size of the initial training set that is formed from the generated population. The limit state function for each selected sample for training is evaluated using a flow-based model which is problem specific and often computationally demanding. Subsequently, a $BART$ model is adaptively trained guided by an active learning model that considers the proximity to the limit state and the uncertainty around the estimate of the limit state value provided by the $BART$ model of the previous iteration. A new stopping criterion for the termination of the adaptive training of the $BART$ model is introduced. This stopping criterion, which is founded on the proposed learning function, assesses the value of information in the remaining set of candidate design samples for its potential to improve the performance of the surrogate model in analyzing network reliability. The performance of the proposed method is evaluated for the estimation of flow network reliability of IEEE 30, 57, 118, and 300-bus power systems. The Alternating Current ($AC$)-based Cascading Failure ($ACCF$) model proposed by Li et al. (2018) is used in this study as the flow-based model in order to estimate power loss. The $ACCF$ model captures the amount of power flow loss as well as branch failures in a power system using $AC$ power flow and optimal power flow analyses. The performance of $ANR$-$BART$ is compared with crude $MCS$, $SS$, passively trained $BART$ and deep neural network ($DNN$). A comprehensive set of choices for the parameters of the $SS$ model is considered as there is no definitive process to select these parameters. It should be noted that such an examination for $SS$ parameters may not be practical as it defeats the purpose of reducing the computational cost. Moreover, the hyperparameters of the passively trained surrogate models are optimized through cross-validation for fair comparison.

The rest of the paper is organized as follows. Section 2 presents the concept of flow network reliability, an introduction to $BART$, and the proposed active learning algorithm. Section 3 explains a model for cascading failure of power systems, which is used to estimate the capacity of power systems exposed to disturbances. In Section 4, the numerical results of flow network reliability analysis for the benchmark power systems are presented. Summary and concluding remarks of this study are presented in Section 5.

## 2. METHODOLOGY
### 2.1. Flow network reliability
A network can be represented by a graph $G = (N, E)$, where $N$ and $E$ denote a set of nodes and a set of edges in this graph, respectively. When the network is exposed to a disturbance, network components (*i.e.,* nodes and edges) may lose their functionality which can cause disruptions in the network performance. Any interruptions in the network flow may pose a significant threat to public safety, regional and national economies, thus, it is important to estimate the reliability of networks prior to the occurrence of hazards in order to avoid or minimize subsequent losses. Flow network reliability enables evaluating the performance of networks in terms of the probability of failing to satisfy an intended performance level considering both aleatoric and epistemic uncertainty sources in networks and their stressors. Reliability analysis of a system is performed using a limit state function $G(x)$. This function determines the state of the system such that $G(x) \leq 0$ represents failure and $G(x) > 0$ indicates survival. The boundary region where the limit state function is equal to zero, $G(x) = 0$, is referred to as the limit state ($LS$), where $x$ is the set of random variables. In flow network reliability, the limit state function can be defined as the difference between the flow capacity and demand of the network. Thus, a failure event occurs when the demand of the network reaches or exceeds its capacity, and the probability of failure can be determined as:

$$P_f = P(G(x) \leq 0) \tag{1}$$

In flow network reliability estimation, the performance of networks is often evaluated using a flow-based analysis, and subsequently the reliability of networks is evaluated for the intended performance level. A few studies (*e.g.,* Guidotti et al., 2017) proposed simplified measures of flow to estimate the performance of networks exposed to a disturbance without performing detailed flow analysis. However, using these simplified measures does not accurately reflect the network performance (Li et al., 2017). In fact, optimal flow analyses need to be performed to properly quantify the performance of networks in the face of extreme events. The optimal network flow problem can be defined as a constrained optimization that minimizes the cost of transporting flow (Bertsekas, 1998). The general form of this constrained minimization problem is as follows:

$$\boldsymbol{\theta}^* = \mathrm{argmin}\, \varphi(\boldsymbol{\theta}) \tag{2}$$

$$\mathrm{s.t.}\ A\boldsymbol{\theta} = b, \tag{3}$$

$$h(\boldsymbol{\theta}) \leq 0, \tag{4}$$

where $\varphi(\boldsymbol{\theta})$ indicates the cost function. The constraints in Eqs. (3) and (4) are the conservation of flow constraints and the link capacity constraints, respectively. The optimization vector $\boldsymbol{\theta}$ indicates unknown variables. In flow network reliability, if a target is set for the demand of the network, the limit state function becomes only a function of the flow capacity of the network. Thus, in the aftermath of a disturbance, the failure event (*i.e.,* $G(x) \leq 0$) can be defined as when the percentage of the flow loss in the network (*i.e.,* the loss in the flow capacity of the network) reaches or exceeds the predefined threshold.

To estimate the flow network reliability using $MCS$, the first step is to generate a Monte Carlo population. Each sample in this population is a vector containing the state of the vulnerable components in the network. Here, the condition of each component is represented using a binary state variable for failure or survival. It is worth noting that the probability distribution representing the condition of a component can be generalized from binary to multinomial. For example, in a transportation network, a damaged bridge can be still partially functional, thus, the condition of edges can be represented by multiple states (Lee et al., 2011). On the other hand, in a power distribution network, the condition of a distribution line is commonly considered as a binary state variable. Although both nodes and edges of a network can fail after a disturbance, without loss of generality, it is assumed that edges are the only vulnerable components that may fail. Subsequently, two adjacent nodes are disconnected if the edge between them is failed. Given the occurrence of an extreme event, the state variable for edge $i$ follows a Bernoulli distribution with failure probability of $p_i$. Thus, each sample consists of a set of zeros and ones that denote the survival and failure of edges, respectively, as follows:

$$x_i = \begin{cases} 1 & \text{if edge } i \text{ is failed} \\ 0 & \text{if edge } i \text{ is survived} \end{cases} \tag{5}$$

where $x_i$ denotes the state of edge $i$, and $\{1, 0\}$ indicates the failure and survival states, respectively. To evaluate each sample, failed edges are removed from the network and a flow-based analysis is performed to estimate the flow capacity of the network. The difference between the flow capacity of the original network (*i.e.*, before removing any edges) and the capacity of the network after the disturbance (*i.e.*, after removing the failed edges) is considered as the loss in the flow capacity of the network. Considering a target flow demand, the network failure (*i.e.*, $G(x) \leq 0$) is defined as when the loss in the capacity reaches or exceeds the predefined demand threshold. Consequently, the network functionality status can be represented as follows:

$$I(x) = \begin{cases} 1 & \text{network failure} \\ 0 & \text{network survival} \end{cases} \tag{6}$$

where $I(x)$ denotes network functionality status for $x$. In the crude $MCS$, all the samples in the population are evaluated and their functionality status is obtained in order to reach an accurate estimate of network failure probability:

$$\hat{u}_{MCS} = \frac{1}{n_{MCS}} \sum_{j=1}^{n_{MCS}} I(x^j) \tag{7}$$

where $\hat{u}_{MCS}$ indicates network failure probability based on the $MCS$ method, and $n_{MCS}$ denotes the number of Monte Carlo simulations (*i.e.*, the number of samples in the population). Consequently, network reliability can be estimated as $\hat{r}_{MCS} = 1 - \hat{u}_{MCS}$.

## 2.2. Bayesian Additive Regression Trees

$BART$ is a highly capable statistical model that leverages the integration of Bayesian probability theories with tree-based machine learning algorithms. A key advantage of $BART$ with respect to other state-of-the-art surrogate models is that it provides an estimate of the uncertainty around the predictions (Hernández et al., 2018). These uncertainties are estimated using a stochastic search based on Markov Chain Monte Carlo ($MCMC$) algorithm. Moreover, the structure of the $BART$ as the sum of regression trees provides high flexibility to conform to highly nonlinear surfaces and the ability to capture complex interactions and additive effects (Chipman et al. 2010; Kapelner & Bleich, 2013; Pratola et al., 2014). Furthermore, $BART$ has shown a high predictive performance when predictors are categorical, binary, and ordinal (Chipman et al., 2010; Tan et al., 2016). As a result, $BART$ offers robust predictive performance for high-dimensional problems with binary variables, and provides the uncertainty associated with the estimated responses (Zamanian et al., 2020b). These features make $BART$ uniquely suited for network reliability analysis, as will be further discussed in the next sections.

The fundamental formulation of $BART$ can be expressed as (Chipman et al. 2010):

$$Y(x) = f(x) + \epsilon, \quad \epsilon \sim N(0, \sigma^2) \tag{8}$$

$$f(x) = \sum_{j=1}^{m} g(x; T_j, M_j) \tag{9}$$

where $x = (x_1, \ldots, x_d)$ denotes the inputs in which $d$ represents the dimension of the predictors. $f(\cdot)$ is the sum of the regression tree models $g(x; T_j, M_j)$. The error term in the model ($\epsilon$) follows a normal distribution with the mean of zero and variance of $\sigma^2$. $BART$ parameterize a single regression tree by a pair of $(T, M)$, where $(T_j, M_j)$ denotes the $j$th regression tree model. An entire tree including the structure and a set of leaf parameters, which represents only a part of the response variability, is formed by an individual function $g(x; T_j, M_j)$ (Zamanian et al., 2020a). $m$ is the number of regression trees. $m$ should be adequately large to yield a high-fidelity $BART$ model. $BART$ uses a group of priors for constructing each single regression tree and a likelihood model for input data in the terminal nodes. Considering the probabilistic representation of $BART$ as:

$$Y(x)|\{(T_j, M_j)\}_{j=1}^m, \sigma^2 \sim N\left(\sum_{j=1}^m g(x; T_j, M_j), \sigma^2\right) \tag{10}$$

the likelihood function is formulated as:

$$L\left(\{(T_j, M_j)\}_{j=1}^m, \sigma^2 | y\right)$$
$$= \frac{1}{(2\pi\sigma^2)^{\frac{n}{2}}} exp\left\{\frac{1}{2\sigma^2}\sum_{i=1}^n \left(y(x^i) - \sum_{j=1}^m g(x^i; T_j, M_j)\right)^2\right\} \tag{11}$$

where $y(x^i)$ is the response of $x^i$ observation and $y$ is:

$$y = (y(x^1), \dots, y(x^n)) \tag{12}$$

The prior helps $BART$ to assign a set of rules to the fit to control the contribution of each single tree to the overall fit. Constraining the extent of contribution of a single tree to the overall fit helps to avoid overfitting. The full prior can be obtained as:

$$\pi\{(T_j, M_j)\}_{j=1}^m, \sigma^2 = \pi(\sigma^2) \prod_{j=1}^m \pi(M_j|T_j)\pi(T_j) \tag{13}$$

where the terminal nodes are parametrized as follows:

$$\pi(M_j|T_j) = \prod_{b=1}^{B_j} \pi(\mu_{jb}) \tag{14}$$

where the prior of $\mu_{jb}$ follows a normal distribution with the mean of zero and variance of $\frac{0.5}{k\sqrt{m}}$. The hyperparameter $k$ controls the prior probability that $E(y|x)$ is constrained in the interval $(y_{min}, y_{max})$ based on a normal distribution (Kapelner & Bleich, 2016). The purpose of $k$ is to follow model regularization such that the leaf parameters are narrowed (shrunk) toward the center of the distribution of the response. The larger values of $k$ results in the smaller values of variance on each leaf parameter. According to the formulation of $BART$, there are four other hyperparameters which define the prior. $\alpha$ and $\beta$ are the base and power hyperparameters in tree prior, respectively, which determine whether a node is nonterminal. In a tree, nodes at depth $d$ are nonterminal with prior probability of $\alpha(1 + d)^\beta$ in which $\alpha \in (0,1)$ and $\beta \in [0, \infty]$ (Chipman et al. 2010; Kapelner & Bleich, 2016). The hyperparameters $q$ and $v$ are also used to specify the prior for the error variance ($\sigma^2$), and are selected such that $\sigma^2 \sim InvGamma(\frac{v}{2}, \lambda\frac{v}{2})$. $v$

represents the prior degrees of freedom and $\lambda$ is specified from the data such that there is a $q^{th}$ quantile ($q\%$ chance) that the $BART$ model will improve in terms of the Root-Mean-Square Error ($RMSE$). $q$ is the quantile of the prior on the error variance (Kapelner & Bleich, 2016). Therefore, as $q$ increases the error variance ($\sigma^2$) decreases.

In this study, $BART$ is fitted using Metropolis-within-Gibbs sampler coupled with $MCMC$ approach, which employs a Bayesian backfitting to generate draws from the proposed posterior distribution (Hastie & Tibshirani, 2000; Kapelner & Bleich, 2013). The derived distribution following burned-in $MCMC$ iterations provides the uncertainty estimates of the posterior samples. Subsequently, Bayesian credible intervals for the conditional expectation function can be obtained for the predicted values. The credible intervals which account for uncertainties of the estimated responses will be used for active learning in the proposed framework.

## 2.3. Adaptive network reliability analysis

In the context of network reliability analysis, several surrogate modeling techniques have been used to estimate failure probability. The training process of these models is passive meaning that the training samples are chosen randomly. Due to the large number of network components, complexity of the network topology, and dependency among the components of a network, passive network reliability methods require a large number of training samples to reach a proper estimate for network failure probability. As mentioned earlier, for flow network reliability, evaluation of training samples can be time-consuming due to the computational demand of a detailed flow-based analysis. In this study, we introduce an active learning method for flow network reliability analysis using $BART$. The proposed method, Adaptive Network Reliability analysis using $BART$ ($ANR$-$BART$) is presented next.

### 2.3.1. $ANR$-$BART$

$ANR$-$BART$ consists of nine stages, which are explained in detail below:

1. *Generate a Monte Carlo population in the design space* ($\Omega$). The size of the design space is obtained based on the number of crude Monte Carlo simulations required to yield accurate estimates of the failure probability. This number, $n_{MCS}$, can be determined as (Echard et al., 2011):

$$n_{MCS} = \frac{1 - u_G}{u_G \times (COV_{u_G})^2} \tag{15}$$

where $u_G$ and $COV_{u_G}$ indicate the failure probability of network and its coefficient of variation, respectively. Considering the fact that prior to performing $ANR$-$BART$, the failure probability is not known, $u_G$ and $COV_{u_G}$ need to be assumed in this step. Therefore, in Eq. (15), $u_G$ and $COV_{u_G}$ are replaced with $\tilde{u}_G$ and $COV_{\tilde{u}_G}$, respectively, where $\tilde{u}_G$ indicates the assumed probability of failure and $COV_{\tilde{u}_G}$ denotes the considered coefficient of variation for $\tilde{u}_G$. This assumption will be checked after the probability of failure is estimated. A population $\Omega$ containing $n_{MCS}$ randomly generated candidate training samples is constructed. This set remains the same in the entire active learning process unless $\tilde{u}_G$ becomes greater than the estimated failure probability, $\hat{u}_G$, in Stage 5.

2. *Select an initial training set* ($S$). The initial training set should be selected randomly as there is no information about the importance of candidate training samples prior to performing the adaptive training process. It is quite important to select a proper size for the initial training set, $S$, since a small size $S$ may not be sufficiently informative and lead to an increase in the number of adaptive iterations, whereas selecting an initial training set with a large size may be computationally inefficient. Most of the studies on active learning algorithms have determined the size of the initial training set based on a rule of thumb (*e.g.,* Bichon et al., 2008; Echard et al., 2011; Pan and Dias, 2017). In this paper, a novel probabilistic criterion is developed for a systematic identification of the size for the initial training set. According to this criterion, the size of $S$ is related to the of probability of having at least $\hbar$ failure samples in $S$. Subsequently, identifying the size of $S$ is facilitated by the fact that each individual response in $S$ follows a Bernoulli distribution. Therefore, their summation, which indicates the total number of failures in $S$, follows a Binomial distribution as:

$$N_{f,s} \sim Bionmial(n_S, \tilde{u}_G) \tag{16}$$

where $N_{f,s}$ is the number of failure samples in the initial training set and $n_S$ represent the number of samples in $S$. Probability of having at least $\hbar$ failure samples in $S$ (*i.e.,* $\Pr(N_{f,s} \geq \hbar)$) can be expressed using the cumulative distribution function of the Binomial distribution as follows:

$$\Pr(N_{f,s} \geq \hbar) = 1 - F(\hbar; n_S, \tilde{u}_G) = 1 - \sum_{i=0}^{\hbar-1} \binom{n_S}{i} (\tilde{u}_G)^i (1-\tilde{u}_G)^{n_S-i} \tag{17}$$

where $F(\hbar; n_S, \tilde{u}_G)$ denotes the cumulative distribution function of $N_{f,s}$. For the case of having at least one failure sample in $S$, $n_s$ can be determined as:

$$n_S = \frac{\log(1 - \Pr(N_{f,s} \geq 1))}{\log(1 - \tilde{u}_G)} \tag{18}$$

where $\Pr(N_{f,s} \geq 1)$ indicates the probability of having at least one failure sample in $S$, which should be defined by the user. In selecting the initial sample size, another important consideration is that the size of the initial training sample set should be larger than the number of input random variables (Srivastava and Meade, 2015). Therefore, the size of $S$ is selected as:

$$n_S = \max\left(\frac{\log(1 - Pr(N_{f,s} \geq 1))}{\log(1 - \tilde{u}_G)}, n_{RV}\right) \tag{19}$$

where $n_{RV}$ denotes the number of input random variables. The randomly selected samples in the initial training set are evaluated using the flow-based model.
3. *Construct an optimized surrogate model (OPT-BART).* $BART$ is trained using the bartMachine package (Kapelner & Bleich, 2016). To achieve the best $BART$ model, the hyperparameters of the model are optimized using a grid search over different combinations of hyperparameters (Kapelner & Bleich, 2016). Subsequently, the selected combination of hyperparameters that results in minimum $RMSE$ is integrated to Eq. (10) to construct the full prior for the $BART$ model. The imposed prior assists in effective regularization of $BART$ to prevent the effect of each individual tree becoming unduly influential (Kapelner & Bleich, 2016; Chipman et al. 2010). The constructed $BART$ model in this stage is referred to as $OPT$-$BART$ since its hyperparameters are optimized. Although it is ideal to determine optimal hyperparameters in every adaptive iteration, the process of search can be time-consuming. The decision about reoptimizing the hyperparameters is taken in Stage 7.
4. *Identify the best next training samples in $\Omega$, $S_{A_i}$, to enhance the performance of the fitted BART model.* In this stage, the fitted $BART$ model is used to evaluate all the samples in $\Omega$ based on a learning function (the proposed learning function is discussed in Section 2.3.2). The learning function is developed based on the capability of $BART$ in estimating the uncertainties around the predictions. First, the fitted $BART$ model is used to determine the expected value of the limit state function and the associated uncertainty for all samples in $\Omega$. Subsequently, the learning function is computed for each individual sample in $\Omega$. The best set of samples in $\Omega$ is selected for the adaptive training stage to enhance the accuracy of the surrogate model in capturing the limit state function. The set of best additional training samples in step $i$ is called $S_{A_i}$.

5. *Evaluate a stopping criterion for the active learning process*. Once the best next samples in $\Omega$ are identified (*i.e.,* $S_{A_i}$), the stopping criterion needs to be evaluated. If the stopping criterion is satisfied, the surrogate model has reached the desired level of accuracy and does not require additional training. The stopping criterion depends on the choice of the learning function, which is discussed in Section 2.3.3.
6. *Update the training set*. In this stage, all the samples in $S_{A_i}$ are evaluated and then added to the existing training set $S$.
7. *Consider reoptimization of hyperparameters of BART*. Since the hyperparameters of the $BART$ model are first optimized based on the initial training set ($S$), these hyperparameters need to be reoptimized as the size of the training data set increases. This process improves the accuracy of the $BART$ model in the active learning process. Reoptimization can be performed based on either the percentage of the population ($\Omega$) used for training or the number of iterations in the active learning process. The latter is used in this study in which the $BART$ model is reoptimized every 50 iterations.
8. *Train the surrogate model adaptively (A-BART)*. The optimal hyperparameters used to construct the $OPT$-$BART$ model are used to adaptively train a new $BART$ model for the updated set $S$. The new model called $A$-$BART$ is then used to evaluate all samples in $\Omega$ and obtain the values of the learning function.
9. *End of ANR-BART*. Once the stopping criterion is met and the estimated network failure probability is greater than the assumed probability of failure in Stage 1, $ANR$-$BART$ stops and the failure probability of the network is determined.

The overall flow of the proposed $ANR$-$BART$ method is shown in Fig. 1.

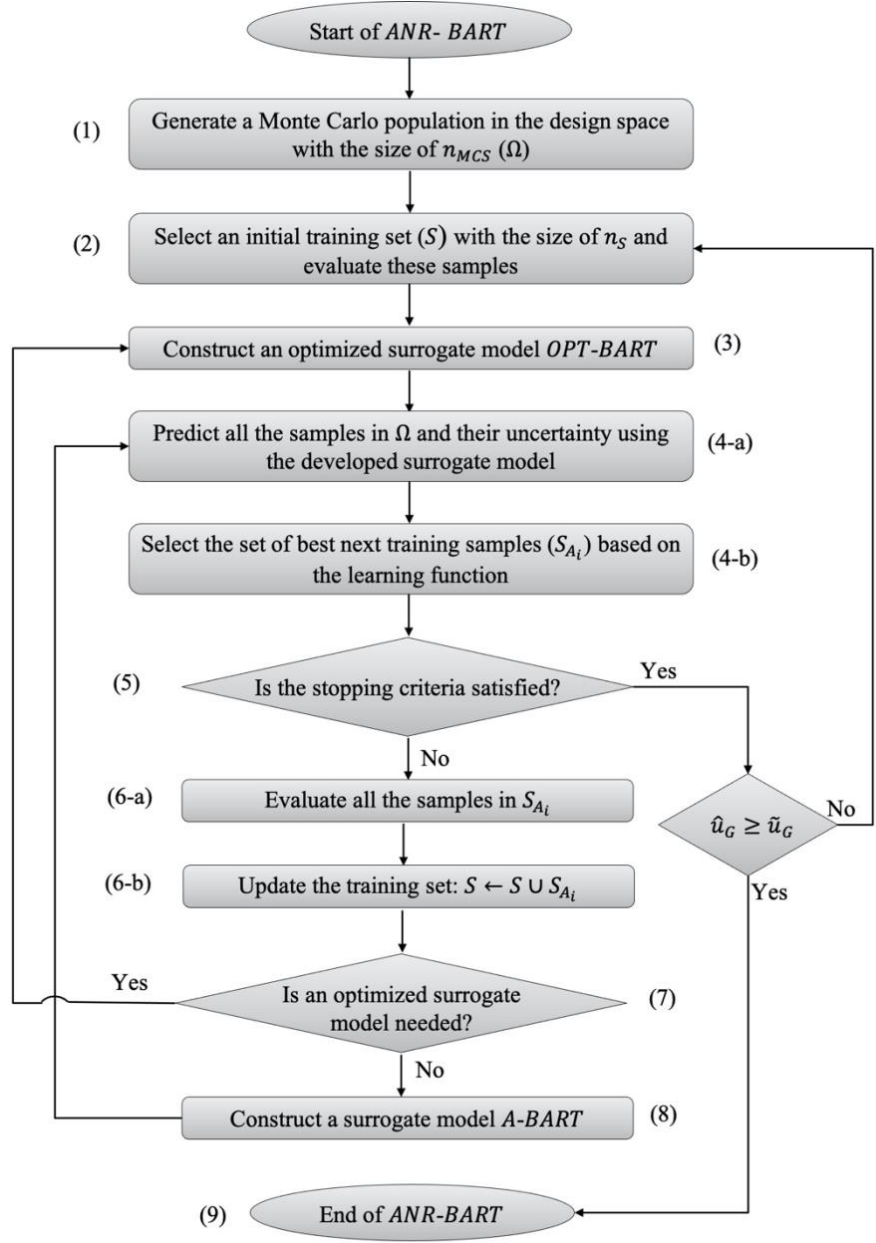

Fig.1 $ANR$-$BART$ algorithm flowchart

The primary feature of $ANR$-$BART$ is that it uses estimates of uncertainty around the predictions and the proximity to the limit state function to identify the most valuable samples among the large set of candidate training samples for the adaptive training of the $BART$ model. This approach affords $ANR$-$BART$ requiring a significantly reduced computational cost especially for reliability analysis of complex, high-dimensional network problems.

### 2.3.2. Learning function
In order to implement $ANR$-$BART$ for network reliability analysis, a learning function is needed to identify the next set of best training samples in each iteration to enhance the performance of the $BART$ model. Here, we propose a learning function for $ANR$-$BART$ called $U'$ based on the concept of $U$ learning function proposed by Echard et al. (2011). Candidate training samples with the highest uncertainty in their estimate of the limit state function have the highest likelihood of being misclassified by the $BART$ model. Moreover, the classification accuracy for candidate training samples that are close to the $LS$ are of high importance for the accuracy of the classification over the entire sample space. Hence, $U'(x)$ is defined as:

$$U'(x) = \frac{|\hat{G}(x)|}{CI_{ub}(x) - CI_{lb}(x)} \tag{20}$$

where $\hat{G}(x)$ represents the estimate of the limit state function. $CI_{ub}(x)$ and $CI_{lb}(x)$ are the upper and lower bound of the credible interval, respectively, for a training sample. The distance between $CI_{ub}$ and $CI_{lb}$ represents the extent of uncertainty in the prediction. Moreover, $|\hat{G}(x)|$ indicates the proximity of the training sample to the $LS$. Based on the formulation of Eq. (20), samples with the smallest values of $U'(x)$ are the best candidate samples for training. Due to the regression-based nature of $BART$, adding one sample at a time to the training sample set for adaptive training will have near to negligible impact on the $BART$ model, and therefore is not efficient. Following a comprehensive analysis, it is found that 20% of the initial sample size is appropriate for the number of training sample in $S_{A_i}$.

### 2.3.3. Stopping criterion

In an active learning algorithm, a stopping criterion is required to determine whether the active learning process can stop. When the stopping criterion is satisfied, it indicates that the surrogate model is sufficiently accurate. This criterion can be defined based on the learning function. As mentioned earlier, in each iteration, $S_{A_i}$ is selected based on $U'(x)$ values such that the selected samples have the highest uncertainty and/or are the closest to $LS$ among all samples in $\Omega$. Therefore, adding these samples with the smallest $U'(x)$ to the training set can improve the surrogate model for estimating the failure probability of the system. As such, the expected value of $U'(x)$ in $S_{A_i}$ can be a representative of the value of the sample set for training the surrogate model. Therefore, the relative difference of the expected values in two consecutive iterations can be used in the stopping criterion to capture the degree of improvement gained by the new training set. Based on this concept, a metric is defined to track the changes in the expected value of $U'(x)$ as follows:

$$\delta_i = \frac{\left| E\left[U'_{S_{A_i}}(x)\right] - E\left[U'_{S_{A_{i-1}}}(x)\right] \right|}{E[U'_{\Omega}(x)]} \tag{21}$$

where $\delta_i$ denotes the value of the stopping criterion in $i$th active learning iteration, and $E\left[U'_{S_{A_i}}(x)\right]$ indicates the expected value of $U'(x)$ for $S_{A_i}$. Similarly, $E\left[U'_{S_{A_{i-1}}}(x)\right]$ is the expected value of $U'(x)$ for the best training samples of $(i-1)$th iteration. $E[U'_{\Omega}(x)]$ is the expected value of $U'(x)$ for the entire population. As recommended by Basudhar and Missoum (2008) and further modified by Pan and Dias (2017), in order to implement a more stable criterion and avoid premature termination of the training process, the data points of $\delta_i$ are fitted by a single-term exponential function as follows:

$$\hat{\delta}_i = Ae^{Bi} \tag{22}$$

where $\hat{\delta}_i$ denotes the estimated value of $\delta_i$. $A$ and $B$ are the parameters of the fitted exponential curve. It is worth noting that $A$ and $B$ may vary as the number of iterations increases (*i.e.,* the number of data points of $\delta_i$ increases). In order for the training process to stop, the maximum of $\hat{\delta}_i$ and $\delta_i$ should be smaller than a target threshold $\varepsilon_1$ (Pan and Dias, 2017). Moreover, the slope of the fitted curve at convergence should be between $-\varepsilon_2$ and zero, where $\varepsilon_2$ is a small positive value (Basudhar and Missoum, 2008; Pan and Dias, 2017). To arrive at a more robust convergence, relative changes in the slope of the fitted curve should reach between $-\varepsilon_2$ and zero. Here, $\varepsilon_1$ and $\varepsilon_2$ are set to 0.002 and 0.01, respectively. The stopping criterion is summarized as:

$$SC_i = \begin{cases} 1 & \text{if } \max(\hat{\delta}_i, \delta_i) < \varepsilon_1 \text{ and } -\varepsilon_2 < \frac{BAe^{B(i-1)} - BAe^{Bi}}{BAe^{Bi}} < 0 \\ 0 & \text{Otherwise} \end{cases} \quad (23)$$

When $SC_i$ becomes one, the adaptive training process stops.

## 3. CASCADING FAILURE MODEL OF ELECTRIC POWER SYSTEMS

Although the proposed method in Section 2 can be used for network reliability assessment of any physical network, this study focuses on power systems. Power outages pose significant economic and social impacts on communities around the world. The increasing reliance of the society on electricity reduces the tolerance for power outages, and consequently highlights the need for reliability assessment of these critical lifeline networks in the face of natural and manmade stressors. Several studies performed reliability assessment of power systems using $MCS$ techniques (*e.g.,* Duenas-Osorio and Hernandez-Fajardo, 2008; Cadini et al., 2017; Li et al., 2018; Faza, 2018). In each Monte Carlo realization, the performance of a power system under failure of the components of the system (*i.e.,* nodes and edges) should be analyzed in order to quantify the power loss in that system.

Analyzing the performance of power systems can be a complex task considering that local failures have the potential to cause overloads elsewhere in the system (Li et al., 2018; Faza, 2018) resulting in cascading failures across the system (Baldick et al., 2008). To simulate power system dynamics, several cascading failure models have been proposed. For the sake of computational efficiency, many of these models are developed based on Direct Current ($DC$) power flow analysis. However, these models are unable to approximate the actual cascading process (Li et al., 2017). A few studies developed cascading failure models based on Alternating Current ($AC$) power flow analysis to eliminate underlying assumptions of $DC$-based models (*e.g.,* Mei et al., 2009; Li et al., 2017; Li et al., 2018). In this study, an $AC$-based Cascading Failure model (called $ACCF$ model) proposed by Li et al. (2017) and improved by Li et al. (2018) is used to assess the performance of power systems under failure event of the system's components. The $ACCF$ model is built on $AC$ optimal power flow ($AC$-$OPF$) analysis. In the remainder of this section, optimal power flow ($OPF$) analysis and $ACCF$ model are briefly introduced.

### 3.1. $AC$ optimal power flow analysis

An optimal network flow problem typically can be treated as a constrained optimization where the objective function is the sum of cost of flow through the edges. The constraints are defined as capacity limits for each edge and flow conservation equations at each node. In power systems, an optimal power flow ($OPF$) problem aims to determine unknown parameters in the system such that they minimize the cost of power generation while satisfying all loads and power flow constraints (Glover et al., 2012). The standard version of $AC$-$OPF$ takes the following form (Zimmerman et al., 2010):

$$\min_X \sum_{i=1}^{n_g} f_P^i(p_g^i) + f_Q^i(q_g^i) \quad (24)$$

$$\text{s.t. } g_P(\Theta, V_m, P_g) = P_{bus}(\Theta, V_m) + P_d - C_g P_g = 0 \quad (25)$$

$$g_Q(\Theta, V_m, Q_g) = Q_{bus}(\Theta, V_m) + Q_d - C_g Q_g = 0 \quad (26)$$

$$h_f(\Theta, V_m) = |F_f(\Theta, V_m)| - F_{max} \leq 0 \quad (27)$$

$$h_t(\Theta, V_m) = |F_t(\Theta, V_m)| - F_{max} \leq 0 \quad (28)$$

$$\theta_i^{ref} \leq \theta^i \leq \theta_i^{ref}, \quad i \in \Gamma_{ref} \quad (29)$$

$$v_m^{i,min} \leq v_m^i \leq v_m^{i,max}, \quad i = 1, 2, \ldots, n_b \quad (30)$$

$$p_g^{i,min} \leq p_g^i \leq p_g^{i,max}, \qquad i = 1, 2, \ldots, n_g \tag{31}$$

$$q_g^{i,min} \leq q_g^i \leq q_g^{i,max}, \qquad i = 1, 2, \ldots, n_g \tag{32}$$

where the optimization vector $X$ includes $n_b \times 1$ ($n_b$ indicates the total number of buses or nodes in the power system) vectors of voltage angels $\Theta$ and magnitudes $V_m$ and the $n_g \times 1$ ($n_g$ denotes the total number of generators in the power system) vectors of generators real and reactive power injections $P_g$ and $Q_g$:

$$X = \begin{bmatrix} \Theta \\ V_m \\ P_g \\ Q_g \end{bmatrix} \tag{33}$$

The objective function in Eq. (24) is a summation of polynomial cost functions of real power injections (*i.e.,* $f_P^i$) and reactive power injections (*i.e.,* $f_Q^i$). The equality constraints in Eqs. (25) and (26) represent a set of $n_b$ nonlinear real power balance constraints and a set of $n_b$ nonlinear reactive power balance constraints, respectively. In these two equations, $C_g$ is a sparse $n_b \times n_g$ generator connection matrix with only zero and one entries. The $(i,j)$th element of this matrix is 1 if generator $j$ is located at bus $i$. The inequality constraints in Eqs. (27) and (28) refer to two sets of $n_l$ branch flow limits ($n_l$ denotes total number of branches or edges in the power system) as nonlinear functions of voltage angels $\Theta$ and magnitudes $V_m$. More specifically, Eqs. (27) and (28) indicates the capacity limits at the *from* end and *to* end of branches, respectively. The flows are commonly apparent power flows that are expressed in megavolt-ampere ($MVA$); however, they can be real power or current flows. $F_{max}$ indicates the vector of flow limits. Eqs. (29) to (32) denote the upper bounds and lower bounds of the variables. More details about $AC$-$OPF$ can be found in Zimmerman et al. (2010).

### 3.2. $AC$-based Cascading Failure model

The $ACCF$ model is used here to capture the real power loss (*i.e.,* shed loads) in power systems after random failure of components. Compared to $DC$-based cascading failure models, which are a tractable relaxation of the desirable $AC$-based models, the $ACCF$ model is more accurate in terms of simulating the power system dynamics and computing the power loss (Li et al., 2017). The $ACCF$ model is built on a number of assumptions, including: (1) the state of a branch is treated as binary: failure or survival states; (2) repair actions are not performed during the cascading failure process; (3) initial component failures are mutually independent; and (4) a failed node will cause failure of the connected branches to that node.

According to the $ACCF$ model, a power system is modeled as a directed network with $n_b$ nodes and $n_l$ branches. Nodes represent power generation stations and substations, and branches represent transmission lines. Nodes can be: (1) both power supply and load; (2) only power supply; (3) only load; (4) only a voltage transformation node (neither power supply nor load). Herein, the first two groups of nodes are referred to as supply nodes and the last two categories are referred to as substation nodes. Each branch in power systems is modeled as a standard $\pi$ model transmission line, with series reactance and resistance and total charging capacitance, in series with an ideal phase shifting transformer. The $ACCF$ model first determines isolated islands formed in a power system after the initial failure of its components (*i.e.,* nodes and branches). Then, the model simulates the cascading failure processes for each isolated island separately. For each island, the following steps are performed:

1. Check whether the island contains any supply node. If the island does not contain supply nodes, all power loads in the island are not satisfied, and the cascading failure simulation of the island stops.
2. If there is only one supply node and no substation nodes in the island (*i.e.,* the isolated bus is a supply node), the satisfied power load of that supply node is the minimum value of the power supply and load.

3. If there are supply node(s) as well as substation nodes in the island, it is necessary to conduct an $AC\text{-}OPF$ of the island with randomly assigning a supply node as the reference node. If the $AC\text{-}OPF$ converges, the total amount of satisfied loads is calculated, and the cascading failure simulation of the island stops.
4. If the $AC\text{-}OPF$ does not converge, a load shedding treatment is applied. For this purpose, an $AC$ power flow analysis is conducted for that island to acquire the load of all branches, and consequently determine the overloaded branches. Then, the branch with the highest level of overload is identified and 5% of the load of all nodes within the radius $R$ of that branch are shed. After 5% load shedding, an $AC\text{-}OPF$ of the island is conducted. If the $AC\text{-}OPF$ converges, the total amount of satisfied loads is calculated, and the cascading failure simulation of the island stops.
5. If the $AC\text{-}OPF$ does not converge, Step 4 is repeated up to 20 times. If the convergence is not reached after 20 iterations, the branch having the highest level of overload will be removed from the system. Removal of this branch may divide the power system into more isolated islands. Thus, all the steps should be repeated until a convergence is achieved or no more overloaded branches exist in the system.

## 4. NUMERICAL EXAMPLES

In this section, the application of $ANR\text{-}BART$ for reliability analysis of IEEE 30, 57, 118, and 300-bus systems is explored for random failure of components. These examples represent realistic power networks with their different dimensions, topologies, and other key characteristics. For instance, the IEEE 30-bus system includes 6 supply nodes, 24 substation nodes, 20 load nodes and 41 branches. On the other hand, the IEEE 300-bus system includes 69 supply nodes, 231 substation nodes, 193 load nodes and 411 branches. Characteristics of these power systems are summarized in Table 1 and their configurations are shown in Fig. 2. The $AC$ power flow and $AC\text{-}OPF$ analyses are carried out using MATPOWER v7.1 (Zimmerman and Murillo-Sánchez, 2020). The $ACCF$ model is used here to capture the real power loss (*i.e.,* shed loads) in the studied power systems after the random failure of components. Although the $ACCF$ model can consider the initial failure of both nodes and branches, in this paper, the branches are the only components subject to initial failure. This assumption has been verified for physical networks subject to hazardous events (Guidotti et al., 2017). A straightforward extension is to allow initial failure of both branches and nodes by constructing an auxiliary network with only branch failures, where each node in the original power system is replaced by two nodes connected by an auxiliary branch.

Table 1. Summary of the studied IEEE bus systems

| Power system | | Number | Total real power (MW) | Total reactive power (MVAr) |
|---|---|---|---|---|
| IEEE 30 | Generators | 6 | 335 | 405.9 |
| | Loads | 20 | 189.2 | 107.2 |
| | Branches | 41 | - | - |
| IEEE 57 | Generators | 7 | 1,975.9 | 699 |
| | Loads | 42 | 1,250.8 | 336.4 |
| | Branches | 80 | - | - |
| IEEE 118 | Generators | 54 | 9,966.2 | 11,777 |
| | Loads | 99 | 4,242 | 1,438 |
| | Branches | 186 | - | - |
| IEEE 300 | Generators | 69 | 32,678.4 | 14,090.2 |
| | Loads | 193 | 23,847.6 | 7,707.6 |
| | Branches | 411 | - | - |

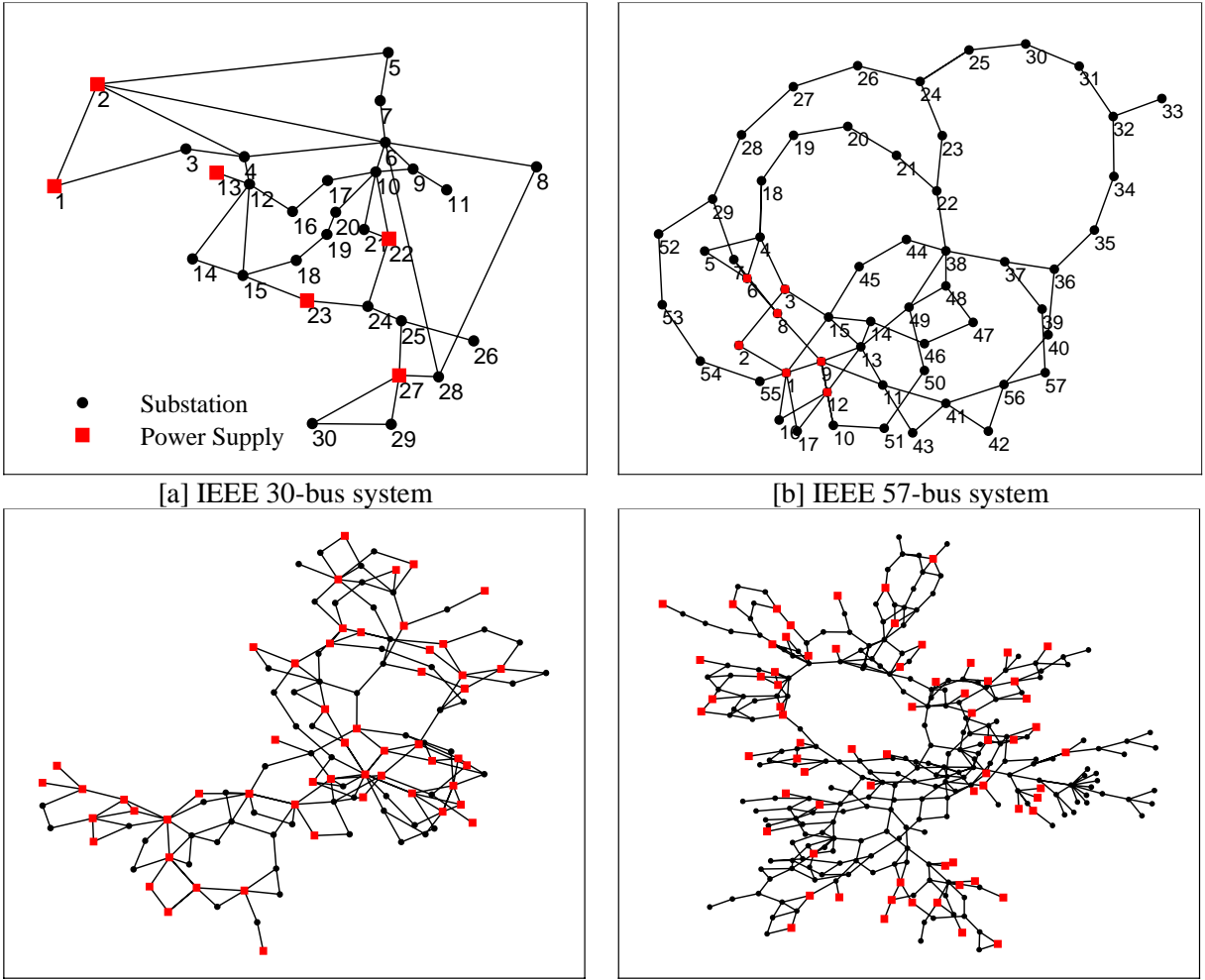

[a] IEEE 30-bus system
[b] IEEE 57-bus system
[c] IEEE 118-bus system
[d] IEEE 300-bus system
Fig. 2 Configuration of the IEEE 30, 57, 118, and 300-bus systems

The performance of $ANR\text{-}BART$ is investigated against the crude $MCS$, $SS$ as well as surrogate model-based reliability analysis methods. $SS$ is a powerful tool that is able to solve a wide range of reliability analysis problems. However, in many cases, $SS$ manifests a large variability in estimating the failure probability (Papadopoulos et al., 2012). Another major limitation of $SS$ in estimating the network failure probability relates to when the failure probability of components varies over time. In fact, in a dynamic network where the failure probability of components varies over time, the crude $MCS$ and $SS$ should be repeated in order to estimate the network failure probability at the time of interest. On the other hand, $ANR\text{-}BART$ addresses this limitation as the $BART$ model in $ANR\text{-}BART$ is trained based on binary inputs (*i.e.,* a set of zeros and ones that denote the survival and failure of branches in the network) and continuous outputs (*i.e.,* loss in the flow capacity of the network). If the failure probability of components varies, a large Monte Carlo population can be generated at the time of interest based on the updated failure probability of components. Then, the binary samples in this population are passed as inputs to the previously trained $BART$ model in order to determine the loss in the flow capacity of the network for each sample. The failure probability of the network can be subsequently estimated using the loss in the flow capacity of the network for all samples.

The basic concept of $SS$ is to express the failure $G(x) \leq 0$ as the intersection of several intermediate failure events $G(x) \leq b_i, i = 1, \ldots, M$, where $b_i$'s are intermediate threshold values and $b_M = 0$. Therefore, for a small probability of failure, $P_f$ is estimated as a product of conditional probabilities that are larger than $P_f$. Consequently, estimation of these conditional probabilities requires fewer number of calls to the limit state compared to the number of calls required by crude $MCS$. In $SS$, the Modified Metropolis algorithm ($MMA$) (Au and Beck, 2001) is used for sampling from the conditional distributions. $MMA$ is an $MCMC$ simulation technique, which is originated from the Metropolis-Hastings algorithm. The choice of proposal distribution in $MMA$ affects the efficiency of Markov Chain samples (Au and Beck, 2001). As recommended by many studies (*e.g.,* Papadopoulos et al., 2012; Breitung, 2019), uniform distribution is adopted here as the proposal distribution. Another major issue with implementing $SS$ is the choice of the intermediate failure events (Au and Beck, 2001; Papadopoulos et al., 2012). In other words, it is difficult to specify the value of $b_i$'s in advance. To alleviate this problem, the $b_i$'s are selected adaptively such that the estimated conditional probabilities are equal to a predefined value $p_0 \in (0, 1)$ that is referred to as the intermediate conditional probability. In each level of this adaptive process, $n_l$ samples are needed to obtain an accurate estimate of the intermediate conditional probability, where $n_l$ should be specified in advance. The choice of both $p_0$ and $n_l$ can affect the efficiency and robustness of $SS$ (Papadopoulos et al., 2012). Several studies recommended implementing $SS$ with an intermediate conditional probability (*i.e.,* $p_0$) within the range of 0.1-0.3 and with the number of samples at each conditional level (*i.e.,* $n_l$) within the range of 500-2000 (Au et al., 2007; Zuev et al., 2012). However, these values are problem specific. For example, Papadopoulos et al. (2012) reached a proper estimate of the failure probability with selecting $n_l$ as high as 10,000, while Li and Cao (2016) selected $n_l$ as low as 300. Due to the lack of a systematic strategy for selecting $p_0$ and $n_l$, 27 combinations of these parameters are investigated in this study. The intermediate conditional probability is considered as $p_0 \in \{0.1, 0.2, 0.5\}$ and the number of samples per conditional level is determined as $n_l \in \{100, 200, 300, 500, 1000, 2000, 3000, 4000, 5000\}$. The implementation of $SS$ follows the algorithm and codes provided by Li and Cao (2016).

In surrogate model-based reliability analysis, $MCS$ is often integrated with a meta-model in order to reduce the computational cost of evaluating the system's performance. In this process, the surrogate model is commonly used as a function approximator to emulate the performance of the system, and consequently to learn the limit state function. In this study, $BART$ and multi-layer perceptron as a subset of deep neural network ($DNN$) methods are used as surrogate models. Hereafter, these approaches are referred to as passive $BART$ and $DNN$, respectively. While $ANR$-$BART$ is integrated with an active learning algorithm which identifies highly uncertain training samples near the limit state in each iteration, the two surrogate model-based reliability analysis methods (*i.e.,* passive $BART$ and $DNN$) randomly select the training samples per iteration. Passive $BART$ is studied to highlight the importance of active learning in flow network analysis, and $DNN$ is investigated here as it is a capable machine learning technique for handling nonlinear problems with high dimensions.

As noted in stage 7 of $ANR$-$BART$ (Section 2.3.1), the hyperparameters of $BART$ are optimized every 50 adaptive iterations. To perform a fair comparison between $ANR$-$BART$ and two passively trained meta-models (*i.e.,* passive $BART$, and $DNN$), the hyperparameters of $BART$ in the passive $BART$ and the hyperparameters of $DNN$ are optimized every 50 iterations. Using the bartMachine package, the optimization of $BART$ model is performed by cross-validation over a grid of hyperparameters (Kapelner & Bleich, 2013). The hyperparameters including $k$, $m$, $q$ and $\nu$ are searched over the ranges provided in Table 2. The values for hyperparameters $\alpha$ and $\beta$ representing the base and power in tree prior, respectively, are adopted from studies by Chipman et al. (2010), Hernández et al. (2018) and McCulloch et al. (2018). For each combination, the out-of-sample Root Mean Squared Error ($RMSE$) is obtained and the lowest one is selected as the best fit named as $OPT$-$BART$. $\hat{Y}(X)$ is obtained as the sample mean of the generated $MCMC$ posterior with 1000 iterations after the first 1000 realizations were burnt. The credible intervals ($CI$) are determined based on post the burned-in $MCMC$ iterations.

Table 2. Selected grid ranges for hyperparameters of $BART$

| Hyperparameter | Vectors | References |
|---|---|---|
| $m$ | {50, 100, 200, 400} | Chipman et al. (2010); Hernández et al. (2018) |
| $k$ | {2, 3, 5} | Chipman et al. (2010); McCulloch et al. (2018) |
| $q$ | {0.85, 0.95} | Chipman et al. (2010); Kapelner & Bleich (2016) |
| $\nu$ | {3, 5, 7} | Chipman et al. (2010); Kapelner & Bleich (2016) |
| $\alpha$ | 0.95 | Chipman et al. (2010) |
| $\beta$ | 2.0 | Chipman et al. (2010) |

Similar to $BART$ models in $ANR$-$BART$ and passive $BART$ approaches, $DNN$ models are trained based on binary inputs (*i.e.,* a set of zeros and ones that denote the survival and failure of branches) and continuous outputs (*i.e.,* real power loss percentage). In this study, Keras package (Chollet, 2015) is used to generate $DNN$ models. $RMSE$ is used as the loss function, and Adam optimization algorithm (Kingma and Ba, 2014) with a learning rate of 0.001 is applied to minimize the loss function. The architecture of $DNN$ surrogate models consists of an input layer with dimension equals to the total number of branches in the power system, a few hidden layers, and an output layer with one dimension. Hyperopt library (Bergstra et al., 2015) is used to select an optimal architecture for the $DNN$ surrogate models, including the total number of hidden layers, their dimensions, and their activation functions. Herein, the total number of hidden layers can vary between four and seven layers where the dimension of each layer can be $2^{n_d}$ with $n_d$ ranging from 1 to 8. The activation function for each layer can be selected between Rectified Linear Unit (ReLU) and Sigmoid functions. In tuning the hyperparameters, a five-fold cross-validation method is applied and the optimal architecture is selected such that the average of the $RMSE$ of the validation sets is minimized.

**4.1. IEEE 30-bus system**
The IEEE 30-bus system topology is shown in Fig. 2[a]. In this case study, the grid contains 41 branches where the state variable for each branch follows a Bernoulli distribution with failure probability of $p = 2^{-3}$. As mentioned earlier, branches are the only vulnerable components that may fail, thus, the total number of random variables is 41. In this paper, the network failure probability is defined as the probability that the real power loss of the network reaches or exceeds a limit. This limit is considered as 30% for this example.

To apply $ANR$-$BART$ for reliability estimation, first, $n_{MCS}$ should be determined. Following Step 1 of $ANR$-$BART$, $n_{MCS}$ is obtained as 40,000 using Eq. (15) with the assumption of $\tilde{u}_G$ equal to 0.01 and $COV_{\tilde{u}_G}$ of 0.05. Subsequently, using Eq. (19) in Step 2, the number of training samples in the initial sample set (*i.e.,* $n_S$) is obtained as 230. In the next step, the $BART$ model is optimized using the hyperparameters defined in Table 2, and the $BART$ model with hyperparameters of $m = 50; k = 2; q = 0.85,$ and $\nu = 5$ is selected as it results in the lowest $RMSE$. The constructed optimal $BART$ model (*i.e., OPT-BART*) is then used to estimate the expected value of the limit state function and the associated uncertainty for the remaining samples in $\Omega$. Consequently, the best next training samples (*i.e.,* $S_{A_1}$) are identified for the active learning process. To calculate $U'(x)$ using Eq. (20), the upper and lower credible intervals as well as the distance of the samples to the $LS$ (*i.e.,* $|\hat{G}(x)|$) are needed. Considering that the stopping criterion is not satisfied in this iteration, the current training set is updated (*i.e.,* $S \cup S_{A_1}$). Following this process, in each adaptive iteration $i$, a set containing the next best training samples (*i.e.,* $S_{A_i}$) is added to the current training set $S$ as follows:

$$Adaptive\ iteration\ 1: S \leftarrow S \cup S_{A_1}$$
$$Adaptive\ iteration\ 2: S \leftarrow S \cup S_{A_2}$$
$$\vdots$$

The size of $S_{A_i}$ is considered as 20% of the size of the initial sample set. Thus, in this example, 46 samples are added to the current training set $S$ per adaptive iteration.

Fig. 3 shows the variation of 90% credible intervals (*i.e.*, $CI_{ub}(x) - CI_{lb}(x)$) and $\hat{G}(x)$ of the entire population for different iterations. It should be noted that the credible interval for each sample indicates the degree of uncertainty around the estimate for that sample. Moreover, $\hat{G}(x)$ of each sample refers to the estimate of the limit state function for that sample. In other words, $\hat{G}(x)$ denotes $LS - \hat{Y}(x)$, where $\hat{Y}(x)$ is the estimated mean of sample $x$ using $BART$ model. Therefore, $\hat{G}(x) \leq 0$ indicates the set of estimated failure samples using $BART$. Prior to the start of the adaptive process, the design space is not well classified and only a few failure samples are estimated (Fig. 3[a]). As the adaptive iteration starts, the samples with higher credible intervals and lower $|\hat{G}(x)|$, which lead to lower values of $U'(x)$, are identified as the best samples to update $S$. According to Fig. 3, as the number of adaptive iterations increases the number of failure samples increases. This observation shows that the active learning algorithm properly assists $BART$ to learn the failure domain. In general, it can be seen in Fig. 3 that the uncertainty around the estimates of samples in $\Omega$ increases for the first few adaptive iterations (Fig. 3[a]-3[c]). However, this uncertainty decreases as the number of iterations increases (Fig. 3[d]-3[f]). This pattern exists since the surrogate model is trained by only a small set of training samples over the first few iterations; thus, it cannot provide proper estimates of mean and uncertainty around the mean for most of the samples in $\Omega$. As the number of training samples increases, the confidence in the estimates is enhanced. After 180 iterations involving 8,510 training samples, the $BART$ model learns the failure domain with high confidence (Fig. 3[f]), which is essential for estimating the failure probability of the power system.

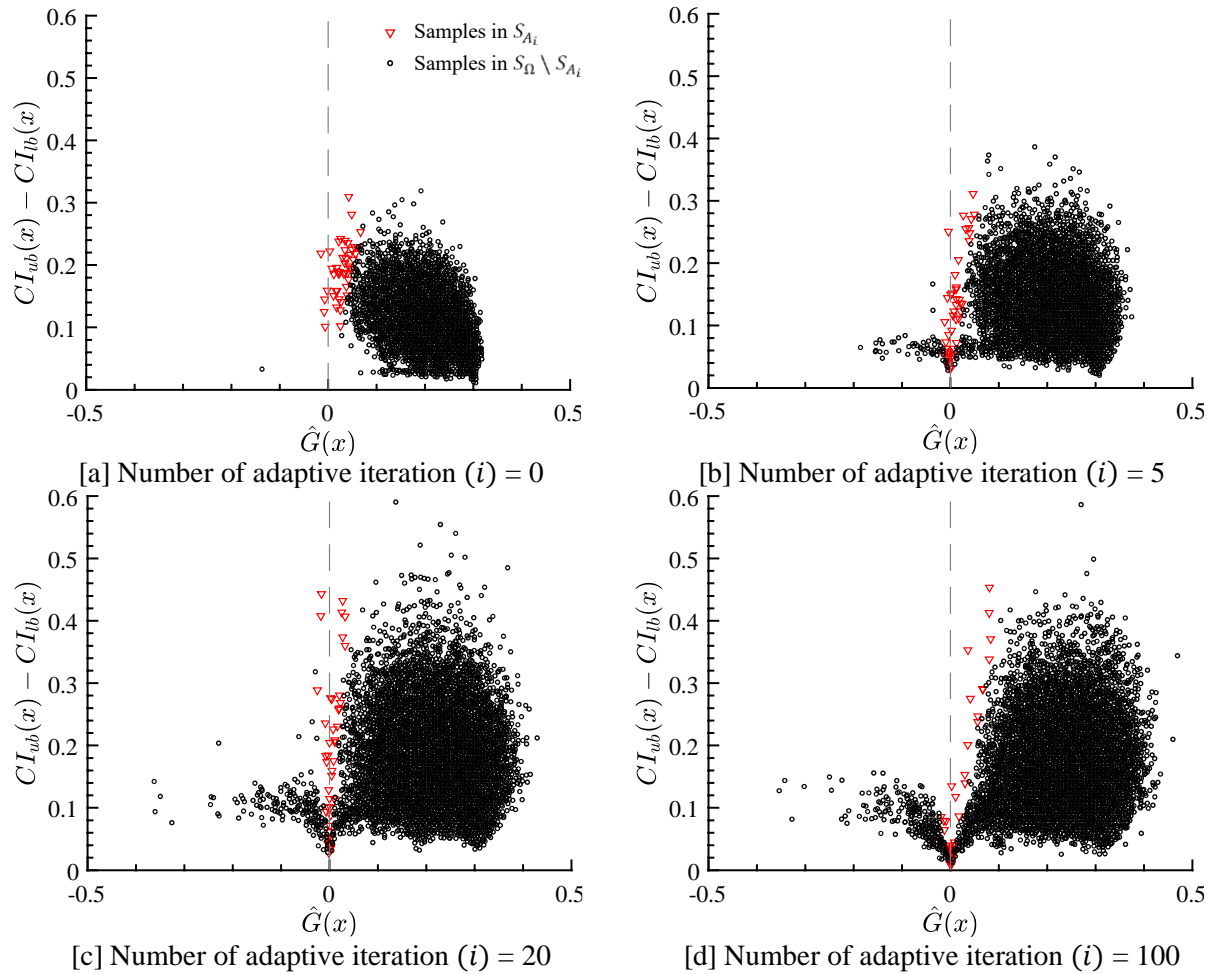

[a] Number of adaptive iteration ($i$) = 0

[b] Number of adaptive iteration ($i$) = 5

[c] Number of adaptive iteration ($i$) = 20

[d] Number of adaptive iteration ($i$) = 100

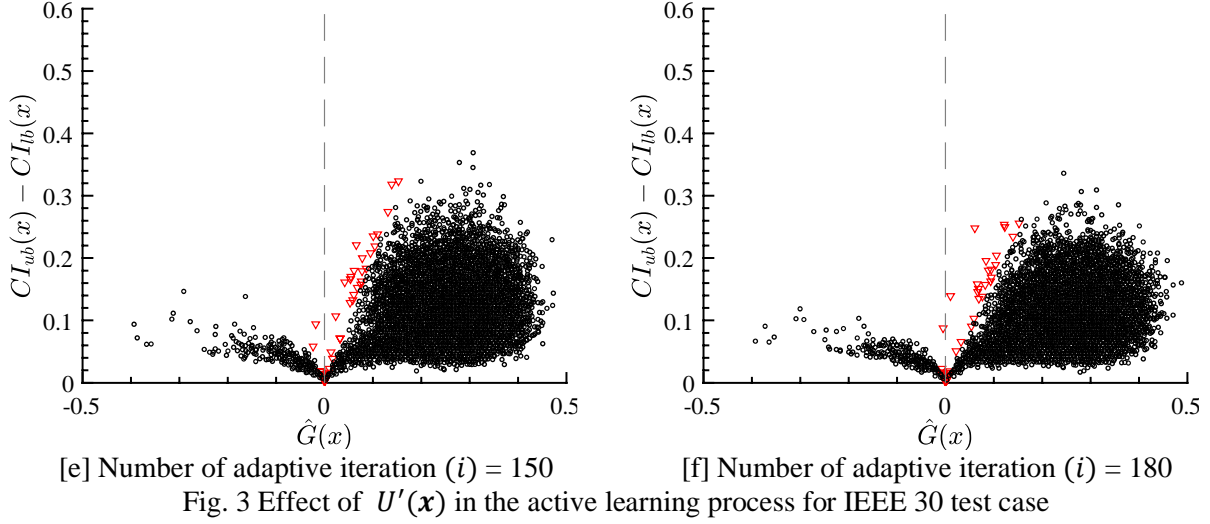

[e] Number of adaptive iteration ($i$) = 150      [f] Number of adaptive iteration ($i$) = 180

Fig. 3 Effect of $U'(x)$ in the active learning process for IEEE 30 test case

The uncertainty around the estimates of samples that contributes to selecting the best next training samples is based on the normality assumption of the error term in the *BART* model (see Eq. (8)). To investigate this assumption – *i.e.*, whether the residuals are normally distributed with the mean of zero and variance of $\sigma^2$ – the empirical cumulative distribution of residuals for the test data for two randomly selected adaptive iterations are presented in Fig 4. According to this figure, the distribution of residuals is in a good agreement with the normality assumption of the error in the *BART* model.

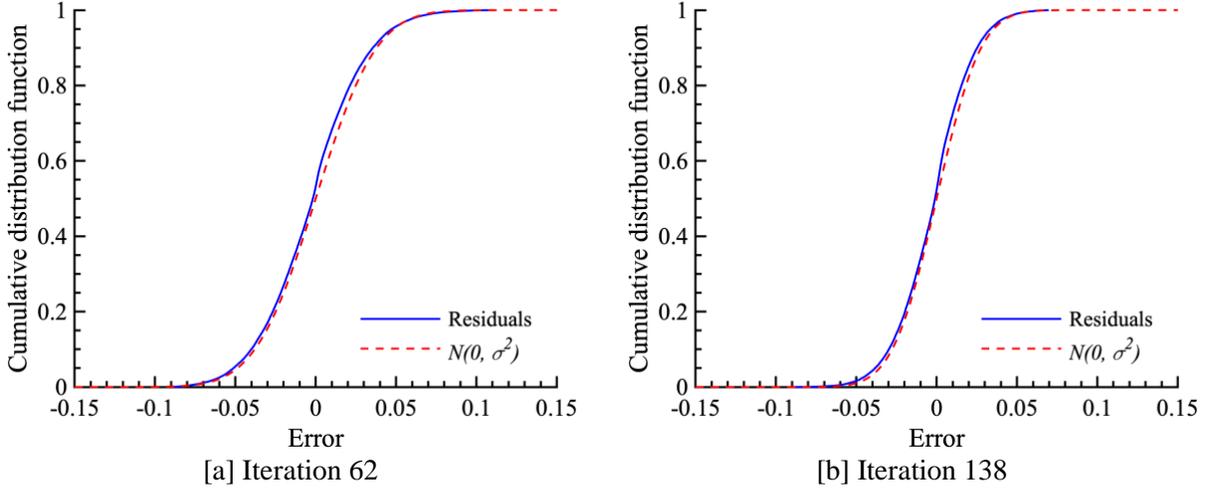

[a] Iteration 62      [b] Iteration 138

Fig. 4 Cumulative distribution function of residuals and the assumed normal error in the *BART* model

Fig. 5[a] and Fig. 5[b] present the expected value of the proposed learning function for the set of best next training samples (*i.e.*, $E\left[U'_{S_{A_i}}(x)\right]$) and the entire population (*i.e.*, $E[U'_{\Omega}(x)]$), respectively, over adaptive iterations. According to Fig. 5, $E[U'(x)]$ is large in the first few iterations. As illustrated in Fig. 3, when the surrogate model is trained by only a small set of training samples, it cannot provide proper estimates of mean and uncertainty around the mean for most of the samples in $\Omega$. More specifically, uncertainty of the estimates is low at early stages, which results in large values of $U'(x)$. After a few iterations, both $E\left[U'_{S_{A_i}}(x)\right]$ and $E[U'_{\Omega}(x)]$ follow an increasing trend implying that the confidence bounds of the *BART* model around its estimates tighten. As explained in Section 2.3.1, hyperparameters of the surrogate model are optimized every 50 iterations. This optimization may yield similar or different

hyperparameters for the $BART$ model. Results in both Fig. 5[a] and Fig. 5[b] indicate that $E[U'(x)]$ suddenly changes in iterations 50 and 150. These sudden changes are due to the updated hyperparameters of the surrogate model. More specifically, in iterations 50 and 100, the hyperparameters for the $BART$ model are $m = 100; k = 2; q = 0.95,$ and $v = 5$, while in iteration 150, the hyperparameters are updated as $m = 200; k = 5; q = 0.85,$ and $v = 5$.

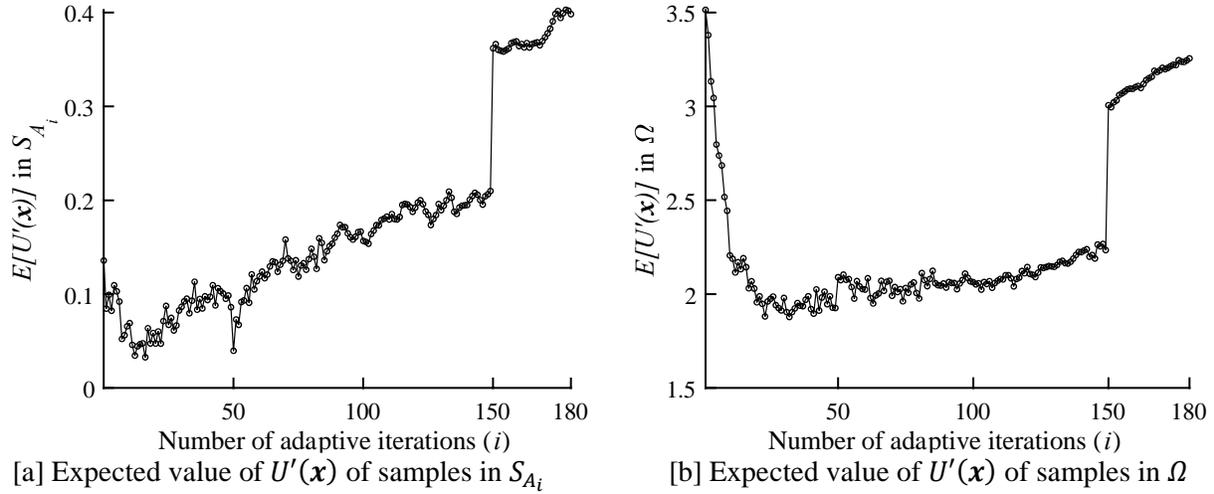

[a] Expected value of $U'(x)$ of samples in $S_{A_i}$  [b] Expected value of $U'(x)$ of samples in $\Omega$

Fig. 5 Evolution of the expected value of $U'(x)$ over the adaptive steps

Fig. 6 shows the convergence curves that are used to evaluate the stopping criterion. In this study, the value of $\varepsilon_1$ and $\varepsilon_2$ for the stopping criterion are set to 0.002 and 0.01, respectively. Considering these thresholds, $ANR\text{-}BART$ is terminated after 180 adaptive iterations. In other words, the actively trained $BART$ reaches an acceptable performance with 8,510 calls to the limit state function, which is about 21% of the required number of simulations for $MCS$.

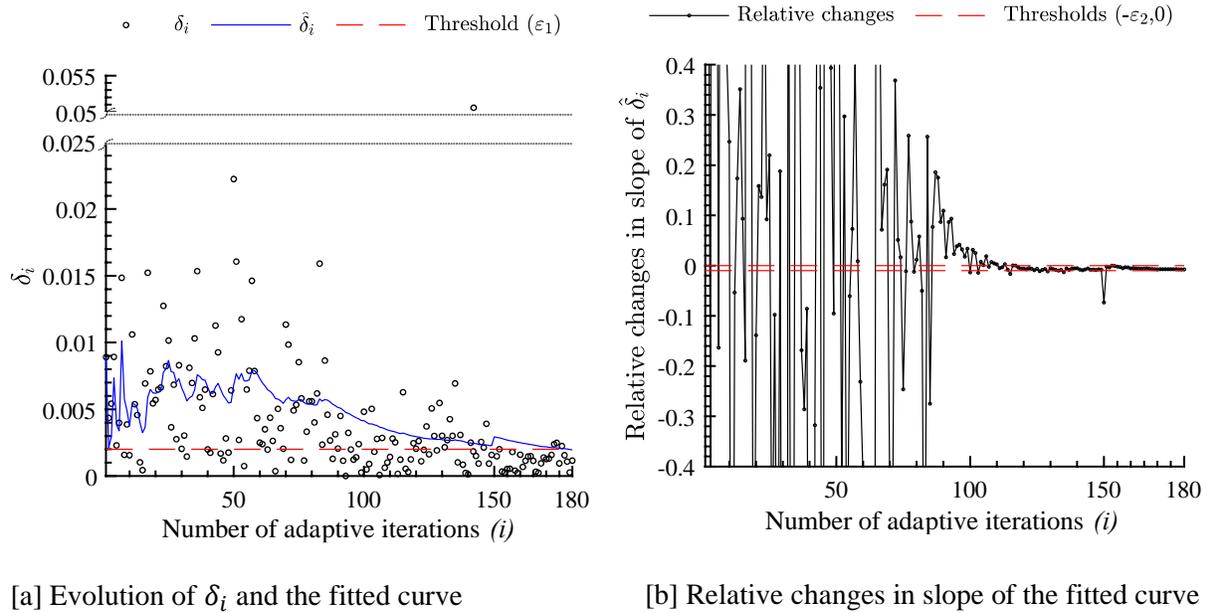

[a] Evolution of $\delta_i$ and the fitted curve  [b] Relative changes in slope of the fitted curve

Fig. 6 Convergence curves of $ANR\text{-}BART$ in IEEE 30 test case

Fig. 7 presents the evolution of the estimated failure probability of the network using $ANR\text{-}BART$ (i.e., $\hat{u}_G$) over the adaptive iterations. The probability of failure of the network using the $MCS$ method (i.e., $u_{MCS}$) is obtained as 0.0143 with around 40,000 calls to the limit state function. In this figure, $\tilde{u}_G$ refers to the assumed failure probability at Stage 1 of $ANR\text{-}BART$. The results indicate that $\hat{u}_G$ determined by $ANR\text{-}BART$ converges to $u_{MCS}$ after 150 iterations; however, due to the conservatively defined stopping criterion, $ANR\text{-}BART$ continues the active learning process for additional iterations.

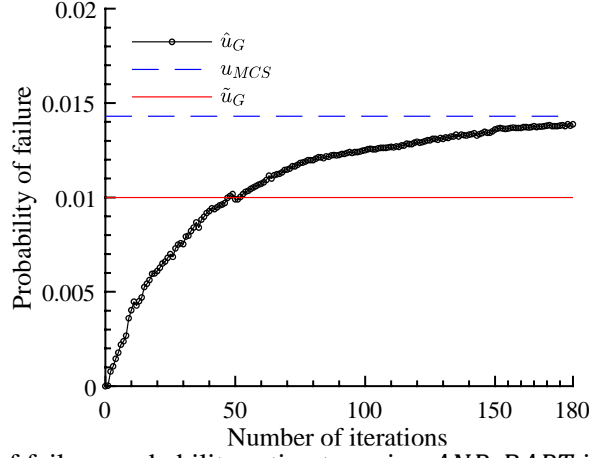

Fig. 7 Evolution of failure probability estimates using $ANR\text{-}BART$ in IEEE 30 test case

To evaluate the performance of the proposed active learning framework, $ANR\text{-}BART$ is compared to $SS$ as well as passively trained $BART$ and $DNN$ models. Neural network is selected as several studies have used these models for reliability analysis of structures and networks (*e.g.,* Hurtado and Alvarez, 2001; Papadopoulos et al., 2012; Nabian and Meidani, 2018). To perform a fair comparison between $ANR\text{-}BART$, passive $BART$, and $DNN$, the same number of training samples used for $ANR\text{-}BART$ are randomly selected from the candidate training set for passive training of the $BART$ and $DNN$ models. Moreover, the passive $BART$ and $DNN$ surrogate models share the same set of training samples in each iteration. The performance of these approaches is compared in terms of the relative error of the estimated failure probability as follows:

$$RE\ (\%) = \frac{|\hat{u}_G - u_{MCS}|}{u_{MCS}} = \left|\frac{\hat{u}_G}{u_{MCS}} - 1\right| \times 100 \tag{34}$$

where $\hat{u}_G$ is the estimated failure probability of the network using these approaches.

As $ANR\text{-}BART$, passive $BART$ and $DNN$ are built on surrogate models, it is possible to track the evolution of the relative errors as the number of calls to the limit state function increases; however, this is not the case for $SS$. Thus, first the change in the relative errors of $ANR\text{-}BART$, passive $BART$ and $DNN$ is shown. Then, the results of $SS$ are presented. Finally, a comparison between all four approaches is provided. Fig. 8 presents the evolution of relative errors over the number of iterations. $ANR\text{-}BART$ reaches the relative error of 3.14% after 180 iterations which represents 8,510 calls to the limit state function (*i.e.,* 21% of the required number of calls for crude $MCS$). However, the relative error rate of passive $BART$ and $DNN$ at this stage is 14.31% and 29.65%, respectively. This result indicates that integrating an active learning algorithm in surrogate-model based reliability analysis can significantly enhance the performance of surrogate models in reaching an accurate estimate of the failure probability by identifying the best training samples. This performance is achieved because in reliability analysis only a region near the limit state function is important and training samples outside that region do not considerably improve the estimates of failure probability. Comparing passive $BART$ and $DNN$ shows that $BART$ is more robust and can reach lower relative errors for this example. Although neural networks have been shown to perform well in nonlinear and high-dimensional problems, they may result in large errors in reliability analysis of high-dimensional systems with highly nonlinear limit state functions. This observation is also compatible with other investigations on reliability analysis of high-dimensional structures using neural networks (*e.g.,*

Papadopoulos et al., 2012). This limitation of neural networks is rooted in their proneness to overfitting, the gradient diffusion problem, and difficulty of parameter tuning (Erhan et al., 2010; Chojaczyk et al., 2015; Wang & Zhang, 2018; Hormozabad and Soto, 2021). More specifically, neural networks need large amounts of training data to provide high-quality results (Aggarwal, 2018), while tree-based models, such as $BART$, are scalable to large problems with smaller number of training data compared to neural networks (Markham et al., 2000; Razi and Athappilly, 2005). Furthermore, the difficulty in the selection of the optimal architecture makes the application of $DNNs$ problem specific, while $BART$ is very robust to hyperparameter selection (Chipman et al., 2010).

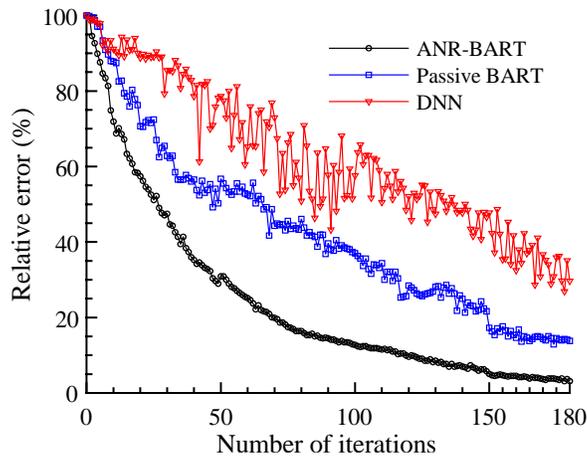

Fig. 8 Relative error of $ANR\text{-}BART$, passive $BART$, and $DNN$ with respect to crude $MCS$ in the IEEE 30-bus system

To further investigate the robustness of $ANR\text{-}BART$, passive $BART$ and $DNN$, the above analyses are repeated nine more times and the mean value, coefficient of variation ($COV$), and maximum value of the relative error with respect to crude $MCS$ as well as the mean value of the total number of calls to the limit state function are reported in Table 3. Moreover, to compare the efficiency, accuracy, and robustness of $SS$ with respect to the proposed method, $SS$ is also performed 10 times and the results are reported in Fig. 9, Fig. 10, and Table 3. To determine the best set of parameters for estimation of the probability of failure of the IEEE 30-bus system using $SS$, 27 combinations of $(p_0, n_l)$ are investigated. For each set of parameters, 10 simulations are performed. According to Fig. 9, as the number of samples per level (*i.e.,* $n_l$) increases, the robustness and accuracy of $SS$ increases. This trend exists because larger values of $n_l$ often results in a larger total number of calls to the limit state function. Although $SS$ reaches small relative errors in some cases, it exhibits large variabilities in the repeated experiments. These observations are compatible with several other investigations *e.g.* Papadopoulos et al. (2012).

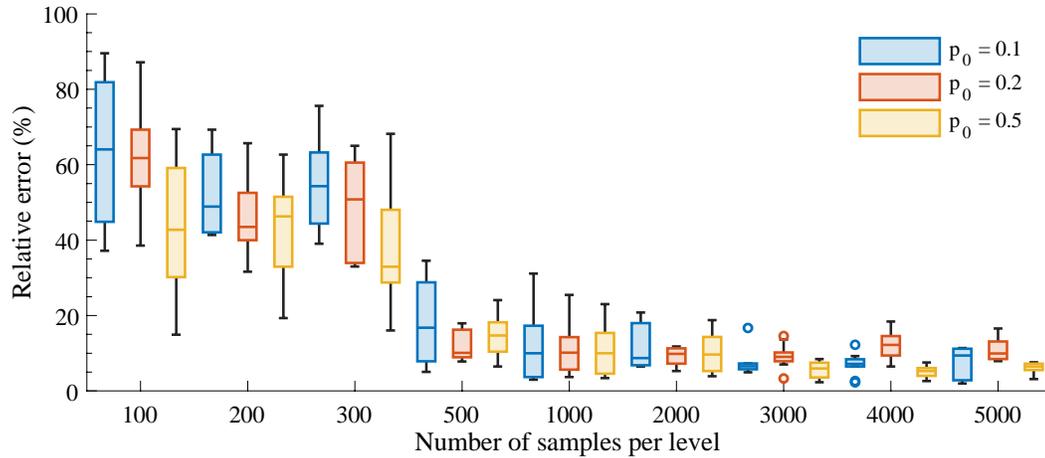

Fig. 9 Relative error of *SS* with respect to crude *MCS* in the IEEE 30-bus system for different combinations of the intermediate conditional probability and number of samples per level

  Fig. 10 presents the mean value of the relative error and the total number of calls to the limit state function of *SS* for different combinations of the intermediate conditional probability ($p_0$) and the number of samples per level ($n_l$). According to this figure, the mean value of relative errors significantly decreases as the number of samples per level passes 300. For $p_0$ of 0.1 and 0.5, changes in the mean value of the relative error become very small as $n_l$ exceeds 2000. According to Fig. 10, the total number of calls to the limit state increases as the intermediate conditional probability increases. This trend is expected because more intermediate levels are needed to reach a proper estimate of the failure probability for larger intermediate conditional probabilities.

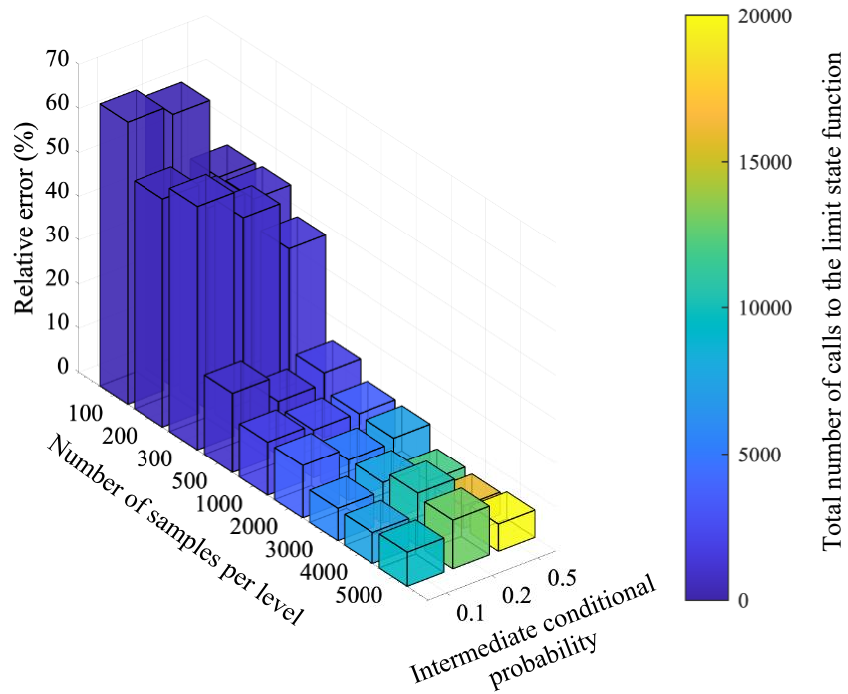

Fig. 10 Mean of the relative error and total number of calls to the limit state function of *SS* for different combinations of the intermediate conditional probability and number of samples per level

Table 3 presents the mean, $COV$, and maximum value of the relative error of $SS$, $ANR\text{-}BART$, passive $BART$ and $DNN$ with respect to crude $MCS$ as well as the mean value of the total number of calls to the limit state function. For $SS$, the table includes the results of the best five combinations of $(p_0, n_l)$ in terms of the mean value of the relative error. It is worth noting that the mean value of the total number of calls to the limit state function for $ANR\text{-}BART$, passive $BART$ and $DNN$ methods are the same in Table 3. This is because passive $BART$ and $DNN$ are terminated in this study when $ANR\text{-}BART$ stops to compare their accuracy and robustness for the same computational cost. According to this table, although $SS$ results in low mean values of the relative error with small number of calls to the limit state function in some cases, using this method for estimating the failure probability of the IEEE 30-bus system leads to large variations. Another limitation of $SS$ is the lack of a priori knowledge about the parameters of the method. As it can be inferred from Fig. 9, Fig. 10, and Table 3, different set of parameters can lead to a significantly different estimate of the failure probability, and consequently a large variation in the relative error. Performing this comprehensive investigation for selection of appropriate parameters for the $SS$ approach is not practical as it defeats the purpose of reducing the computational cost of reliability analysis. Table 3 also shows that $ANR\text{-}BART$ leads to a significantly higher accuracy and robustness in the estimation of the probability of failure compared to other methods, while requiring a reduced number of total calls to the limit state function.

Table 3. Mean value and $COV$ of the relative error and the total number of calls to the limit state function of $SS$, $ANR\text{-}BART$, passive $BART$ and $DNN$ for the IEEE 30-bus system

| Method | | Relative error (%) | | | Mean value of total number of calls to the limit state function |
|---|---|---|---|---|---|
| | | Mean | $COV$ | Maximum | |
| Subset simulation $(p_0, n_l)$ | (0.1, 3000) | 7.33 | 0.46 | 16.70 | 5,700 |
| | (0.1, 4000) | 6.99 | 0.42 | 12.22 | 7,600 |
| | (0.5, 3000) | 5.61 | 0.39 | 8.42 | 12,000 |
| | (0.5, 4000) | 5.05 | 0.33 | 7.56 | 16,000 |
| | (0.5, 5000) | 6.15 | 0.23 | 7.59 | 20,000 |
| $DNN$ | | 31.45 | 0.29 | 43.57 | 8,639 |
| Passive $BART$ | | 19.35 | 0.18 | 23.73 | 8,639 |
| $ANR\text{-}BART$ | | 4.12 | 0.11 | 4.76 | 8,639 |

### 4.2. IEEE 57-bus system

The IEEE 57-bus test case includes 80 branches, therefore the total number of input random variables is 80. The state variable for each branch follows a Bernoulli distribution with failure probability of $p = 2^{-3}$. More details about this grid are presented in Table 1. Given the high computational demand of finding appropriate parameters of $SS$ and the relatively high variation in the estimates of reliability by $SS$, for the case study of IEEE 57-bus system and the remaining examples in this paper, the comparisons are made between $ANR\text{-}BART$ and passive $BART$ and $DNN$ with respect to crude $MCS$. In this example, a failure defined as the event where the real power loss reaches or exceeds 10%. Similar to the previous example, the value of $\varepsilon_1$ and $\varepsilon_2$ for the stopping criterion are respectively set to 0.002 and 0.01. Moreover, $n_{MCS}$ is determined as 40,000 with the assumption that $\tilde{u}_G$ is equal to 0.01 and with $COV_{\tilde{u}_G}$ of 0.05. Subsequently, using Eq. (18), the number of training samples in the initial sample set (*i.e.,* $n_S$) is obtained as 230. Considering the limit state of 0.1, the probability of failure of the network using the $MCS$ method is obtained as 0.013 with 40,000 calls to the limit state function. It is worth noting that the time complexity of the $ACCF$ model does not linearly increase with the size of the gird and it highly depends on the topology of the network as well as the number of substations and power supplies. For example, estimating the failure probability of the IEEE 57-bus test using 40,000 Monte Carlo simulations takes over than a week using

MATPOWER on a personal computer equipped with an Intel Core i7-6700 CPU with a core clock of 3.40 GHz and 16 GB of RAM. However, the same analysis for the IEEE 118-bus system requires slightly over a day.

Fig. 11[a] presents the performance of $ANR\text{-}BART$ in terms of the relative error compared to the passive methods. Fig. 11[b] also compares the estimated flow network reliability using $ANR\text{-}BART$ with $u_{MCS}$ and the assumed failure probability at Stage 1 of the proposed framework.

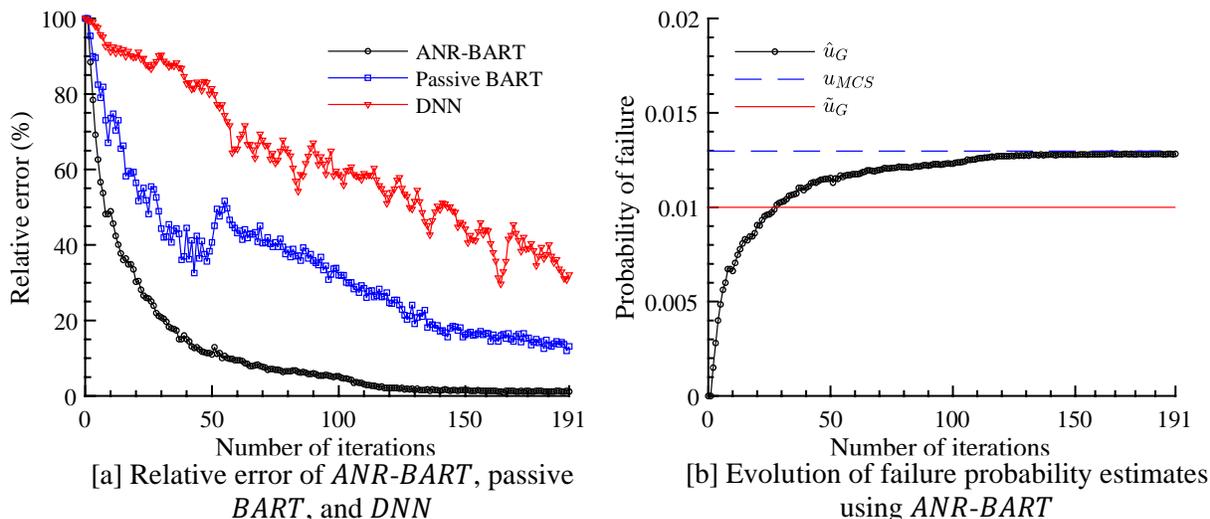

[a] Relative error of $ANR\text{-}BART$, passive $BART$, and $DNN$

[b] Evolution of failure probability estimates using $ANR\text{-}BART$

Fig. 11 The performance of $ANR\text{-}BART$ for the IEEE 57 test case

It is found that using $ANR\text{-}BART$, the relative error reaches 1.16% after 191 iterations with 9,016 calls out of 40,000 required samples for $MCS$. However, the relative error for the same sample size using passive $BART$ and $DNN$ is 13.1% and 30.76%, respectively (Fig 11[a]). This result indicates that the proposed method is able to accurately estimate and quickly converge to the failure probability of the network by identifying the most informative samples via the active learning process. As seen in Fig. 11[a], the relative error of $ANR\text{-}BART$ reaches about 5% after 98 iterations, which indicates a proper estimate of the network failure probability with only 4,738 calls out of 40,000 initial candidate samples. However, due to the conservatism built into the stopping criterion, $ANR\text{-}BART$ continues the active learning process and the stopping criterion is satisfied after 191 iterations. This observation indicates that improving the stopping criterion has the potential to further enhance the performance of the adaptive network reliability analysis.

### 4.3. IEEE 118-bus system

This grid contains 186 branches which indicates that there are 186 random variables in this power flow network reliability analysis problem. Similar to the previous examples, the state variable for each branch follows a Bernoulli distribution with failure probability of $p = 2^{-3}$. Here, the failure is defined as the event where the real power loss reaches or exceeds 10%. The value of $\varepsilon_1$ and $\varepsilon_2$ for the stopping criterion are set to $0.002$ and $0.01$, respectively. Moreover, $\tilde{u}_G$ and $COV_{\tilde{u}_G}$ are set to $0.01$ and $0.05$, respectively. Considering the limit state of $0.1$, the probability of failure of the network using the $MCS$ method is obtained as $0.0149$ with 40,000 calls to the limit state function.

Fig. 12 presents a comparison between $ANR\text{-}BART$, passive $BART$, and $DNN$ in terms of the relative error. $ANR\text{-}BART$ estimates the failure probability of the network with a relative error of 4.72% after 203 iterations with 9,568 calls out of 40,000 required samples for $MCS$. As observed in Fig. 12, $ANR\text{-}BART$ yields accurate estimates of the failure probability with significantly fewer number of calls to the limit state function compared to $MCS$. This steep drop in the number of calls for generating reliable surrogate models substantially increases the efficiency of flow network reliability analysis especially for cases with expensive simulations. However, the relative error for the same sample size using passive $BART$ and $DNN$ is 38.06% and 36.47%, respectively. The results also point to the superior performance of $ANR\text{-}BART$ compared to passive methods as the number of random variables increases.

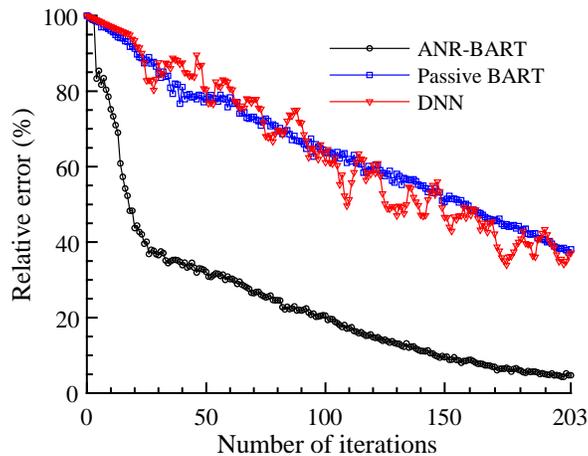

Fig. 12 Relative error of $ANR\text{-}BART$, passive $BART$, and $DNN$ with respect to $MCS$ in the IEEE 118-bus system

### 4.4. IEEE 300-bus system

IEEE 300-bus system contains 411 branches (*i.e.,* 411 random variables) is considered as the most challenging example in this study. Similar to the previous examples, the state variable for each branch follows a Bernoulli distribution with failure probability of $p = 2^{-3}$. Here, the failure is defined as the event where the real power loss reaches or exceeds 50%. The value of $\varepsilon_1$ and $\varepsilon_2$ for the stopping criterion are set to 0.002 and 0.01, respectively. Moreover, $\tilde{u}_G$ and $COV_{\tilde{u}_G}$ are set to 0.01 and 0.05, respectively. Considering the limit state of 0.5, the probability of failure of the network using the $MCS$ method is obtained as 0.0136 with 40,000 calls to the limit state function. Subsequently, using Eq. (18), the number of training samples in the initial sample sets is determined as 230. However, as explained in Section 2.3.1, the size of the initial sample set is selected as the maximum of the value obtained using Eq. (18) and the number of input random variables. Thus, in this example, 411 is selected as the size of the initial sample set. Similar to other examples, the size of $S_{A_i}$ is considered as 20% of the initial sample set size. Therefore, in this example, 82 samples are added to current training set $S$ per adaptive iteration.

    The relative error of $ANR\text{-}BART$ and the studied passive methods are presented in Fig. 13. $ANR\text{-}BART$ estimates the failure probability of the network with 4.43% error after 125 iterations with 10,661 calls to the limit state. However, the relative error for the same sample size using passive $BART$ and $DNN$ is 29.89% and 36.59%, respectively. This observation indicates the capability of $ANR\text{-}BART$ in handling high-dimensional and highly nonlinear problems. Comparing the results of the IEEE 300-bus system and previous case studies reveals that as the size of the network increases, the required number of calls to the limit state for $ANR\text{-}BART$ does not change significantly. More specifically, the number of random variables in IEEE 300-bus system is more than 10 times of the number of random variables in IEEE 30-bus system; however, the total number of calls to the limit state in the former case study only increase by about 5%. In addition to the size of networks, differences in the topology of the case studies as well as the number of substations and power supplies significantly affect the complexity of the physics and other operational relations and constraints that control network operations. Despite all these different sources of complexities, $ANR\text{-}BART$ is able to estimate the network failure probability with less than 5% error in all cases. These results highlight the capability of $ANR\text{-}BART$ in reliability analysis of high-dimensional networks with different characteristics. It is worth noting that in flow network reliability, the performance of networks is often evaluated using a flow-based analysis that requires solving optimal flow analysis ($OPF$). Since $OPF$ is a constrained optimization problem, it can be very computationally demanding if the nonlinearity of the

objective function and constraints increases or the number of the involved variables rises. In these cases, the advantage of using *ANR-BART* in flow network reliability analysis becomes even more evident.

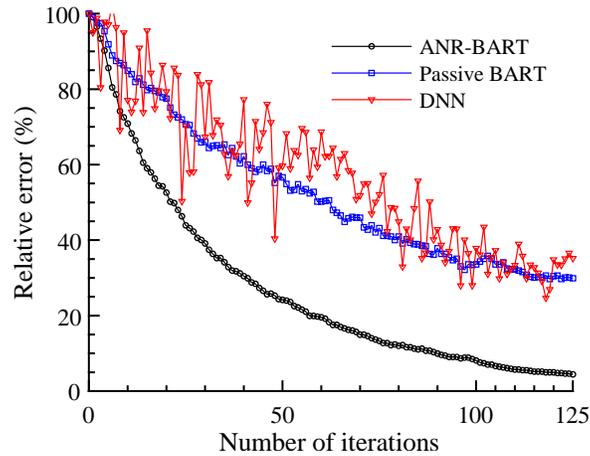

Fig. 13 Relative error of *ANR-BART*, passive *BART*, and *DNN* with respect to *MCS* in the IEEE 300-bus system

## 5. CONCLUSIONS

This paper proposes a robust machine learning approach for conducting efficient and accurate flow network reliability analysis. This approach is called Adaptive Network Reliability analysis using Bayesian Additive Regression Trees (*ANR-BART*), which integrates *BART* and Monte Carlo simulation (*MCS*) via an active learning algorithm. This is the first study that devises an approach for adaptive training of surrogate models for network reliability analysis. In the proposed approach, an active learning function is introduced to identify the most valuable training samples out of the large set of candidate training samples to adaptively train a *BART* model. This function evaluates the importance of each sample based on the best estimate of the dependent variables and the associated credible interval that are provided by the *BART* model of the previous iteration over the space of predictor variables. In this process, the higher the uncertainty in the response of a sample and the closer its proximity to the limit state, the higher the value of the sample for training the *BART* model. The performance of *ANR-BART* is investigated for four benchmark power systems including IEEE 30, 57, 118, and 300-bus systems. Comparing the results of these four case studies highlights the capability of *ANR-BART* in reliability analysis of high-dimensional networks with different properties. This paper also compares the performance of *ANR-BART* with respect to subset simulation (*SS*) as well as surrogate model-based reliability analysis methods where deep neural networks (*DNN*) and *BART* are passively trained. The presented results indicate that *ANR-BART* is an accurate, robust, and significantly more efficient method compared to *SS* and the passive surrogate-based methods for flow network reliability analysis.

## 6. ACKNOWLEDGMENTS
This research has been partly funded by the U.S. National Science Foundation (NSF) through awards CMMI-1635569, 1762918, and 2000156. In addition, this work is supported by an allocation of resources for computing time from Ohio Supercomputer Center. These supports are greatly appreciated.